\documentclass[10pt,twocolumn,letterpaper]{article}

\usepackage[pagenumbers]{cvpr} 

\usepackage[dvipsnames]{xcolor}

\definecolor{cvprblue}{rgb}{0.21,0.49,0.74}
\usepackage[pagebackref,breaklinks,colorlinks,citecolor=cvprblue]{hyperref}
\usepackage{graphicx}
\usepackage{caption}
\usepackage{pgfplots}
\usepackage[linesnumbered,ruled,vlined]{algorithm2e}

\def\paperID{*****} 
\def\confName{CVPR}
\def\confYear{2024}

\begin{document}

\title{HOP: Heterogeneous Topology-based Multimodal Entanglement \\for Co-Speech Gesture Generation}


\author{
Hongye Cheng\footnotemark[1] \textsuperscript{ 1} \quad 
Tianyu Wang\footnotemark[1] \textsuperscript{ 2} \quad
Guangsi Shi\footnotemark[2] \textsuperscript{ 3} \quad
Zexing Zhao\textsuperscript{4} \quad 
Yanwei Fu\footnotemark[2] \textsuperscript{ 5} \\
\textsuperscript{1}College of Mechanical and Electronic Engineering, Northwest A\&F University \\
\textsuperscript{2}Institute of Science and Technology for Brain-Inspired Intelligence, Fudan University \\
\textsuperscript{3}Department of Chemical and Biological Engineering, Faculty of Engineering, Monash University \\
\textsuperscript{4}College of Information Engineering, Northwest A\&F University \quad
\textsuperscript{5}School of Data Science, Fudan University \\
\tt\small {
    guangsi.shi@monash.edu
    \quad yanweifu@fudan.edu.cn
}
}

\twocolumn[{
\maketitle
\begin{center}
    \captionsetup{type=figure}
    \includegraphics[width=\textwidth]{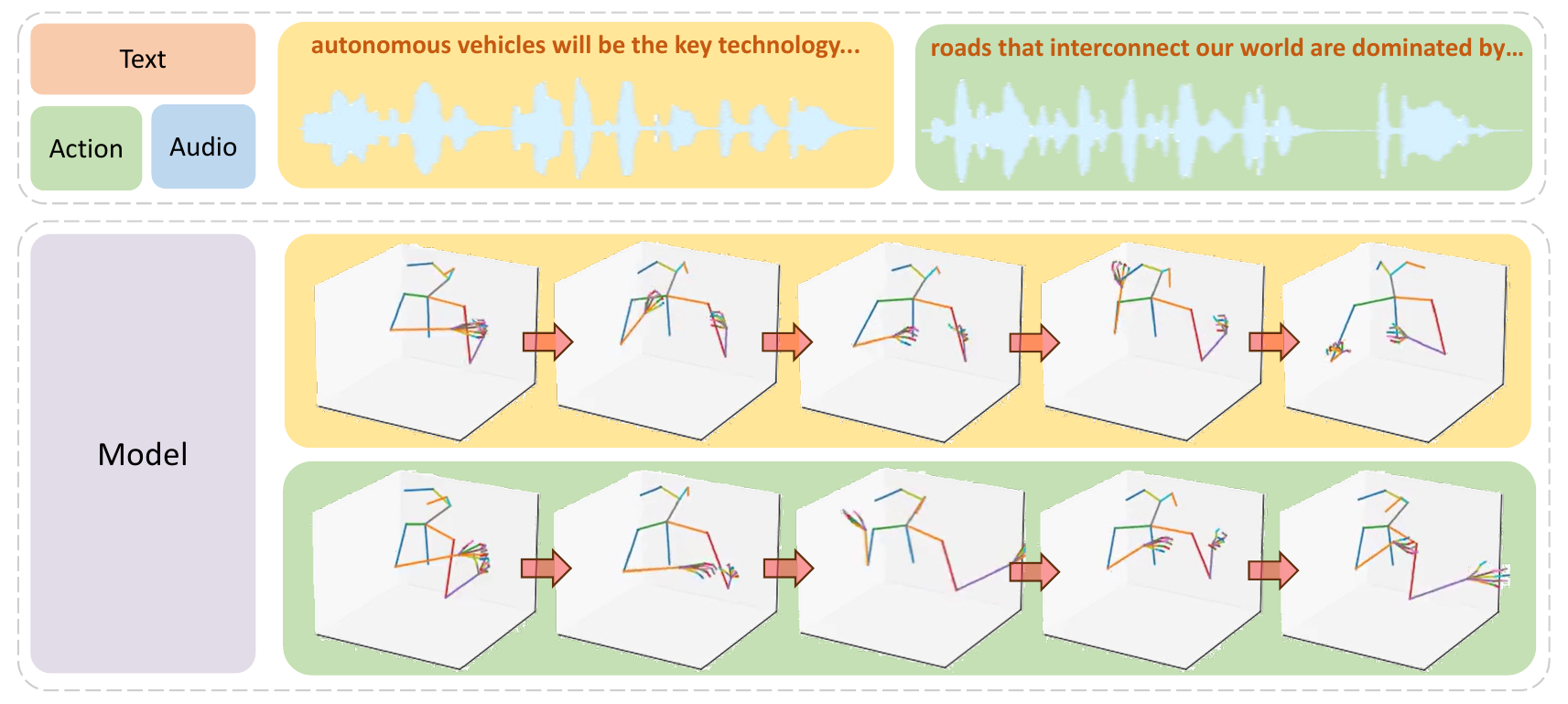}
    \vspace{-0.25in}
    \captionof{figure}{\textbf{HOP:} We propose a topology-based heterogeneous multimodal model that integrates features from audio, text, and action, accounting for their inherent heterogeneity through cross-modality adaptation. The model achieves superior performance on both the TED-Expressive dataset (first row) and the TED dataset (second row), generating gestures that align with the semantics and rhythmic qualities of the speech, as well as the motion characteristics of the real speaker.}
\end{center}
}]
\renewcommand{\thefootnote}{\fnsymbol{footnote}}
\footnotetext[1]{Equal contribution.}
\footnotetext[2]{Corresponding authors.}

\begin{abstract}

Co-speech gestures are crucial non-verbal cues that enhance speech clarity and expressiveness in human communication, which have attracted increasing attention in multimodal research. While the existing methods have made strides in gesture accuracy, challenges remain in generating diverse and coherent gestures, as most approaches assume independence among multimodal inputs and lack explicit modeling of their interactions. In this work, we propose a novel multimodal learning method named HOP for co-speech gesture generation that captures the heterogeneous entanglement between gesture motion, audio rhythm, and text semantics, enabling the generation of coordinated gestures. By leveraging spatiotemporal graph modeling, we achieve the alignment of audio and action. Moreover, to enhance modality coherence, we build the audio-text semantic representation based on a reprogramming module, which is beneficial for cross-modality adaptation. Our approach enables the trimodal system to learn each other's features and represent them in the form of topological entanglement. Extensive experiments demonstrate that HOP achieves state-of-the-art performance, offering more natural and expressive co-speech gesture generation. More information, codes, and demos are available here: \href{https://star-uu-wang.github.io/HOP/}{https://star-uu-wang.github.io/HOP/}.
 
\end{abstract}
    
\section{Introduction}
\label{sec:intro}
\hspace{1em}Co-speech gestures provide crucial non-verbal cues that enhance the clarity and expressiveness of speech. Automatically generating such gestures in virtual avatars and embodied AI agents~\cite{wolfert2022review, liu2021speech, nyatsanga2023comprehensive} has garnered significant attention, as it holds the potential to enhance the realism and interactivity of human-machine interaction.

This task has undergone significant advancements over the years, evolving from early rule-based expert models~\cite{kopp2004synthesizing, wagner2014gesture} to probabilistic approaches~\cite{kipp2007towards, levine2010gesture, neff2008gesture} and, more recently, to deep learning-based methods~\cite{cui2019deep, liu2022audio, li2021audio2gestures, bhattacharya2021speech2affectivegestures}. These developments have led to continual improvements in performance metrics, with deep learning approaches, in particular, achieving notable results in gesture generation tasks. Despite progress in the accuracy of gesture actions, challenges remain in generating gestures that are both diverse and coherent~\cite{kucherenko2023genea, yoon2022genea}.

Most current deep learning methods leverage multimodal approaches~\cite{ramachandram2017deep} to integrate information from multiple domains, such as text, audio, and gesture actions. These multimodal approaches often employ contrastive learning~\cite{yuan2021multimodal}, latent space fusion~\cite{piechocki2023multimodal}, or cross-attention mechanisms~\cite{nagrani2021attention, wei2020multi} to facilitate interaction across modalities. However, such methods typically involve mapping different modalities into a latent space through separate encoder models, with the underlying assumption that these modalities are fully decoupled and independent.

Contrary to this assumption, the evidence in Figure \ref{tsne} suggests that the interactions among modalities are inherently interdependent. When humans communicate, spoken language and gestures are not isolated; rather, they are intertwined, with verbal expression influencing gesture patterns and vice versa. This natural intermodal relationship plays a crucial role in ensuring that gesture generation is coherent and contextually aligned with the corresponding speech~\cite{liu2022learning}. Simple multimodal fusion methods may fail to capture this interdependence, potentially compromising the coherence of generated gestures, especially in scenarios where body movements are essential for enhancing information transmission and achieving a high level of correlation among text, speech, and gestures~\cite{escalera2017challenges}. 

In order to overcome the challenges above, we propose a novel multimodal {entanglement} approach to co-speech gesture generation that explicitly models the topological relationships between action, audio, and text. {Unlike multimodal fusion, where individual modalities are encoded separately and subsequently fused, multimodal entanglement emphasizes the interrelationships among modalities by embedding these dependencies directly into the modal transformations. 
We argue that audio, which inherently encode both the rhythm of gestures and the meaning of the text, provide a natural bridge to align gesture actions and textual information. By leveraging this insight, our approach utilizes audio rhythm as the key to connecting and aligning the information in gesture and text modalities.
 
Our method builds on recent advances in reprogramming techniques~\cite{chen2024model, jin2023time}, which align sequential data from different modalities. In our framework, we align audio rhythm and textual semantics to drive gesture generation. First, we extract audio rhythm features using Mel spectrograms~\cite{sheng2019high}, which capture both the dynamics and frequency of the audio. These features are then integrated into gesture motion by a spatiotemporal graph neural network~\cite{wu2019graph}, allowing us to model the spatial-temporal dependencies of gesture. Finally, we fuse the gesture and text representations and use a Generative Adversarial Network (GAN)~\cite{goodfellow2020generative} to generate realistic gestures.

We introduce a novel multimodal entanglement method to model the topological relationships between action, audio, and text in the context of co-speech gesture generation. Extensive experiments on public datasets demonstrate that our method outperforms state-of-the-art methods across multiple evaluation metrics, including Fréchet Gesture Distance (FGD), Beat Consistency (BC), and gesture diversity, establishing new benchmarks for the task. To summarize, our main contributions are three-fold: 

\begin{itemize}
    
\item{
We propose the novel multimodal framework that explicitly models the topological relationship between gesture motion, audio rhythm, and text semantics for co-speech gesture generation.}
\item{
We introduce a novel approach that leverages reprogramming techniques to align audio rhythm with text semantics, using Mel spectrograms and a spatiotemporal network to capture gesture motion features.}
\item{
Our method achieves state-of-the-art performance on public datasets, demonstrating superior results in FGD, BC, and diversity, establishing a new benchmark for co-speech gesture generation.}
\end{itemize}
\section{Related Work}
\label{sec:rela_works}

\noindent \textbf{Co-speech Gesture Generation}.
Co-speech gesture generation, which aims to synchronize gestures with spoken audio, has garnered significant attention for applications in human-agent interaction~\cite{nyatsanga2023comprehensive, wolfert2022review}. Initially, rule-based methods~\cite{marsella2013virtual, poggi2005greta, wagner2014gesture} defined gesture-audio mappings through expert input, achieving high-quality but rigid outputs. Early work has attempted to combine visemes~\cite{owens1985visemes, kalberer2003visual} and co-speech gestures, aiming to enhance the expressiveness of virtual avatars. Leveraging diverse datasets~\cite{liu2022beat, yoon2020speech, yoon2019robots, zhu2023taming}, deep learning approaches have integrated multiple modalities to enhance the realism of generated gestures, which have proven useful for generating identity-preserving outputs and show strong potential for generalizing to flexible motion sequences. Frameworks using hierarchical and adversarial training~\cite{liu2022learning, yoon2020speech, liu2022beat} have shown promise for joint feature learning, proving effective in enhancing the realism of generated motions. EMAGE~\cite{liu2024emage} explores the application of masked representation learning in this field, employing masked gesture reconstruction to decode pre-trained facial and body latent features. Recently, diffusion-based methods~\cite{zhu2023taming, mughal2024convofusion, yang2023diffusestylegesture} have emerged as a powerful alternative, treating gesture generation as a stochastic process. Our approach aims to enhance the consistency, realism, and smoothness of co-speech gesture generation, establishing a unified model through topology-based multimodal fusion.

\noindent \textbf{Spatial-Temporal Graph Modeling}.
Spatial-temporal graphs are  powerful tools in numerous research domains and industrial applications, including physics simulation, traffic forecasting, and time series anomaly detection. GNSTODE~\cite{shi2024towards}, for instance, introduces a learning-based simulation framework that generates particle dynamics using spatial-temporal graph neural networks. GraphWavenet~\cite{wu2019graph} combines graph structures with wavelet transformations to model spatial-temporal dependencies effectively. Similarly, TSTGNN~\cite{shi2024deep} employs a transformer-based heterogeneous spatiotemporal graph model for enhanced geographical traffic forecasting. GRASS~\cite{liu2023graph} dynamically identifies mode-switching behaviors within time series data, which is critical for capturing complex temporal patterns. In contrast, some anomaly detection frameworks~\cite{su2019robust} are noted to lack explicit modeling of pairwise interdependencies, which can reduce their efficacy in detecting intricate anomalies. For skeleton-based action recognition, MST-GCN~\cite{chen2021multi} introduces a method that decomposes local graph convolution into sub-graph convolutions, creating a hierarchical residual architecture to improve model interpretability and performance. All of these works illustrate the versatility and evolving capabilities of spatial-temporal graph models across diverse applications.

\noindent \textbf{Cross-modality Adaptation}.
Multimodal fusion is crucial for generating realistic gestures aligned with speech and text in co-speech gesture generation. These techniques focus on combining information from various modalities to generate gestures that closely capture the nuances of speech~\cite{nyatsanga2023comprehensive, kim2023mpe4g}. Recently, cross-attention mechanisms~\cite{wang2024cross} and contrastive learning~\cite{teshima2023act2g} have been applied to multimodal fusion, enhancing the dynamic alignment of gesture generation. However, the general fusion may face limitations in adapting to significant differences between input modalities, making it challenging to achieve seamless modality switching and alignment. To address this, cross-modality adaption methods have been developed, introducing shared representation spaces or domain adaptation techniques~\cite{yin2023survey} to enable effective mapping across modalities. Cross-modality adaptation focuses on mapping information effectively between modalities, such as audio and gesture, without extensive reconfiguration. Recent work has explored transferring knowledge from large pre-trained models in NLP and CV through techniques like multimodal fine-tuning and model reprogramming~\cite{chen2024model, yang2021voice2series}. In our approach, we introduce a reprogramming module and specifically constructed cross-modality adaptation module for audio-text and audio-action integration, making joint representation learning more effective and unlocking new potential for powerful multimodal fusion.
\section{Methodology}

\hspace{1em}In this section, we present our approach to leveraging multimodal data with inherent heterogeneity from the speech dataset in the co-speech gesture generation task. Our goal is to generate gestures that not only convey the semantic content intended by the speaker but also align with the speaker's rhythmic delivery. The overall framework of the model is shown in Figure~\ref{framework}.  So we  defines topological relationships in Sec.~\ref{subsec:topological_representation_learning}. Section~\ref{subsec:Audio-Text} introduces the reprogramming module for gesture generation. We integrates audio with action data using spatio-temporal graphs as discussed in Sec.~\ref{subsec:Audio-Action}. We finally  outlines the gesture generator and training objectives in Sec.~\ref{subsec:Generation}.


\begin{figure*}[htb]
	\centering
	\includegraphics[width=1.02\textwidth]{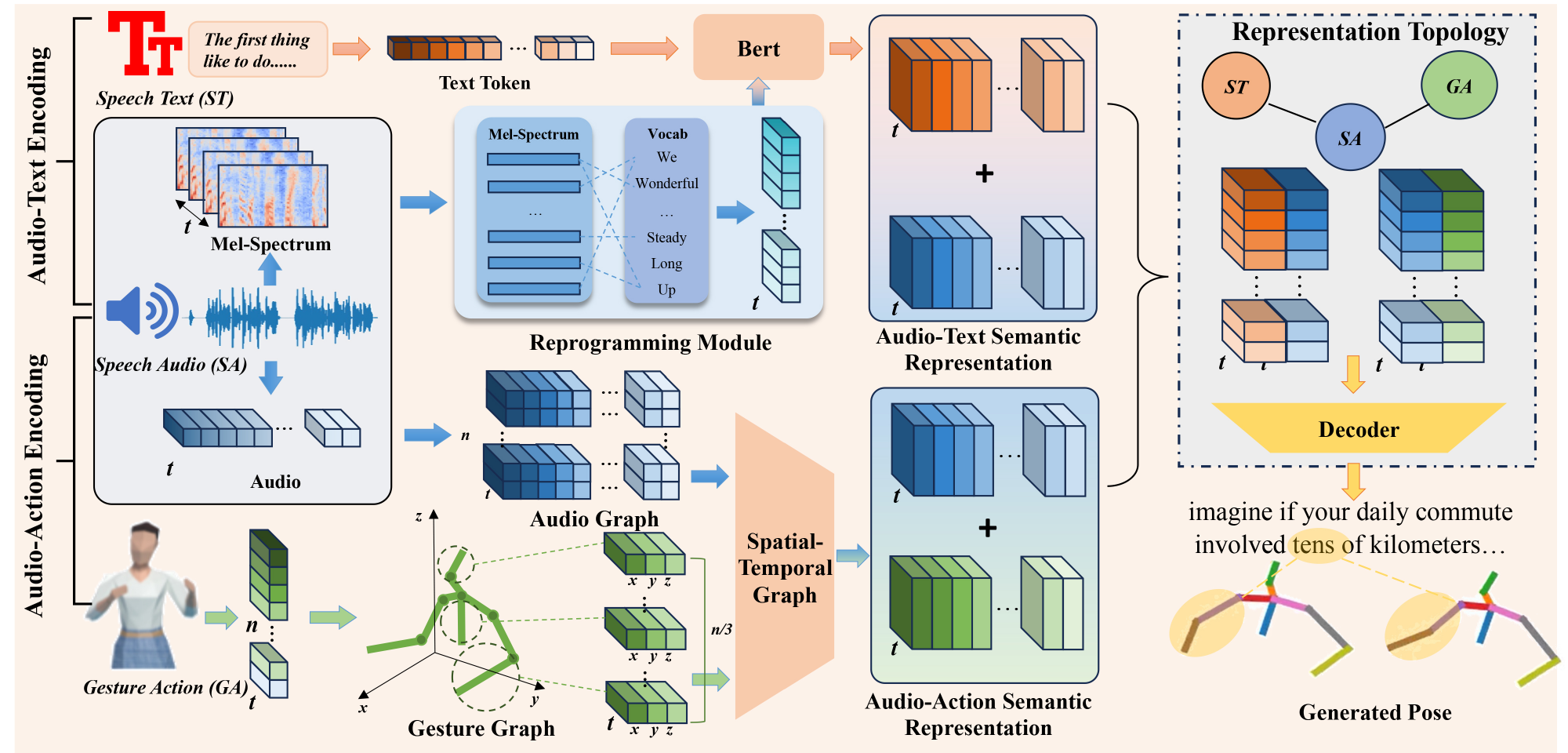}
    \vspace{-0.25in}
	\caption{\textbf{Overview of the proposed framework for multimodal gesture generation with heterogeneous topology entanglement.} Given the input text of speech and the Mel-Spectrum obtained through audio preprocessing, we treat audio sequences as a bridge, linking text sequences and action sequences with distinct topologies. For the connection between text and audio, we apply a reprogramming layer to align data from these different modalities, utilizing a language model to extract embedded semantic information. To link action and audio, we employ the Graph-WaveNet approach to separately extract action and audio features. The entangled multimodal representations are then fed into the gesture generator through topological fusion, resulting in the generation of co-speech gestures.}
\label{framework}
\vspace{-0.2in}
\end{figure*}

\subsection{Topological Representation Learning}
\label{subsec:topological_representation_learning}

\noindent \textbf{Topological Mutimodal Entanglement.} In the context of co-speech gesture generation, we consider three main modalities: \textbf{text($X_t$)}, \textbf{audio($X_{aud}$)}, \textbf{action($X_{act}$)}. A key challenge in this task is to effectively capture the intricate entanglement between multiple modalities. The concept of Topological Multimodal Entanglement (TME) offers a novel approach to modeling the relationships. Our goal is to perform cross-modality adaptation between these modalities, specifically between the \textbf{audio-text} and \textbf{audio-action} representation, as illustrated in the example in Figure~\ref{tsne}. This enables a unified, context-aware generation of gestures that align with both the linguistic content (text), acoustic features (audio), and physical motion (action). Each modality is first encoded through respective encoding functions $f$:
\vspace{-0.1in}
\begin{equation}
    h_t = f_t(X_t), h_{aud} = f_{aud}(X_{aud}), h_{act} = f_{act}(X_{act})
\end{equation}

Then to entangle the modalities, we introduce a cross-modality adaptation layer, where the audio modality is used to refine both the text and action representations. Specifically, the representations $h_t$ and $h_{act}$ are jointly adapted with $h_{aud}$ to ensure alignment:
\vspace{-0.1in}
\begin{equation}
    h_{t-aud} = g_t(h_t, h_{aud}), h_{act-aud} = g_{act}(h_{act}, h_{aud})
\end{equation}

The functions $g_t$ and $g_{act}$ are responsible for aligning the textual and action representations with the audio, ensuring that the speech characteristics influence both the semantic content and the gesture generation.

A key aspect of TME is  topological entanglement, we introduce $f_{tme}$ that integrates these adapted representations:
\begin{equation}
    h_{tme} = f_{tme}(h_{t-aud}, h_{act-aud}, h_{aud})
\end{equation}

\begin{figure}[htb]
	\centering
	\includegraphics[width=0.45\textwidth]{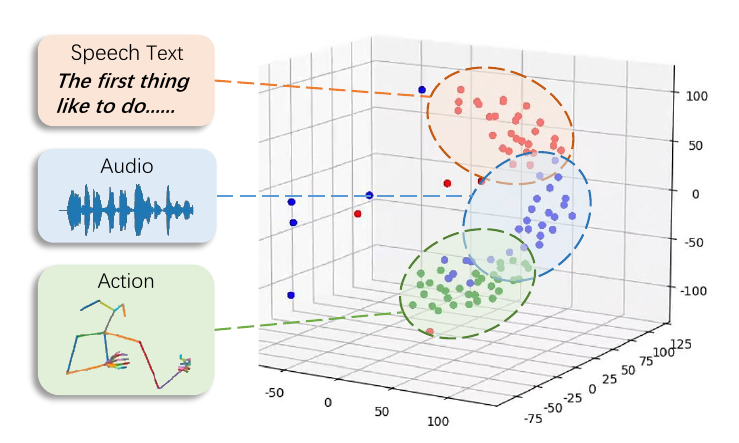}
    \vspace{-0.15in}
	\caption{\textbf{Heterogeneous entanglement of multimodal data.} We use red, blue, and green shading to denote text data, audio data, and action data, respectively. While text and action exhibit significant heterogeneity, audio serves as a direct mediator between the two, establishing a path of connectivity that facilitates the full utilization of multimodal data for gesture generation.}
\label{tsne}
\vspace{-0.15in}
\end{figure}

We explore the topological relationships across multimodal data, particularly between audio and other modalities, and use topological representation learning to capture these interactions. By incorporating this topological information, we aim to improve feature extraction and generate more realistic, coherent co-speech gestures.


\subsection{Audio-Text Cross-modality Adaptation}
\label{subsec:Audio-Text}

\hspace{1em}Previous co-speech gesture generation methods~\cite{yoon2020speech,liu2022learning} primarily employed temporal convolutional networks to extract semantic features from text data. However, research~\cite{loehr2012temporal} has indicated that distinct semantic characteristics within audio can lead to varied granularity in human pose movements. Consequently, we leverage the advanced reasoning capabilities of large language models (LLMs) to extract deeper semantic information from multimodal data. Nonetheless, a fundamental challenge remains: audio data cannot be losslessly represented in natural language, complicating the process of effectively inputting multimodal data into LLMs. To address this, we introduce a reprogramming module ~\cite{jin2023time}, applied for the first time in gesture generation, to facilitate multimodal data fusion.

In this approach, we reprogram the Mel-spectrogram features of audio into a format compatible with the input space of the LLM, enabling it to process and reason about audio data. We denote the Mel-spectrogram features series at time step \textit{t} as $\mathbf{M}^{(t)}\in\mathbb{R}^{1\times T}$. However, due to the inherent limitations of natural language in fully capturing audio content, we propose to utilize pre-trained word embeddings $\mathbf{E} \in \mathbb{R}^{V \times D}$ to reprogram audio data, where $V$ represents the vocabulary size. Given that we cannot determine precisely which vocabulary elements will capture the nuances of audio, a large vocabulary would lead to an excessively broad reprogramming space, thereby consuming significant computational resources. To mitigate this, we map the large vocabulary to a smaller feature space using a linear layer, denoted by $\mathbf{E'} \in \mathbb{R}^{V' \times D}$, where $V^{\prime} \ll V$.

For alignment between audio information and vocabulary, we propose to a multi-head cross-attention layer. For each head $n = {1, \dots, N}$, we define the query matrices $\mathbf{Q}_n = \mathbf{M}\mathbf{W}_n^Q$, key matrices $\mathbf{K}_n = \mathbf{E}^{\prime} \mathbf{W}_n^K$, and value matrices $\mathbf{V}_n = \mathbf{E}^{\prime}\mathbf{W}_n^V$, where $\mathbf{W}_n^Q \in \mathbb{R}^{d_m \times d}$ and $\mathbf{W}_n^K, \mathbf{W}_n^V \in \mathbb{R}^{D \times d}$. Here, $D$ denotes the hidden dimension of the text encoder, and $d = \lfloor \frac{d_m}{N} \rfloor$, where $d_m$ is the dimension of the Mel frequency energy features. The reprogramming operation for time-series patches within each attention head is thus defined as: 

\begin{align} 
\mathbf{Z}_(w,r)^{(1:T)} = E_a(\hat{w}_{1:T},w_{1:T}) \end{align}

\begin{equation}
\hat{w}_{1:T} = Linear(\operatorname{Softmax}\left(\frac{\mathbf{Q} \mathbf{K}^{\top}}{\sqrt{d}}\right) \mathbf{V})
\end{equation}

By aggregating the outputs $\mathbf{Y}_k^{(i)} \in \mathbb{R}^{P \times d}$ across all heads, we obtain $\mathbf{Y}^{(i)} \in \mathbb{R}^{P \times d_m}$, which is then linearly projected to align with the backbone model's hidden dimensions, yielding $\mathbf{O}^{(i)} \in \mathbb{R}^{P \times D}$.

\subsection{Audio-Action Cross-modality Adaptation}
\label{subsec:Audio-Action}

\hspace{1em}While successfully extracting semantic information from text and audio to generate text-aligned gestures, the resulting motions often appear mechanical and subdued. This limitation stems from inadequate learning of the rhythmic characteristics in audio and the dynamic qualities of gesture motion, which reduces the naturalness and expressiveness of generated actions. Conventional methods ~\cite{yoon2020speech,liu2022learning} typically input only the initial four frames from the ground truth into a multi-layer bidirectional GRU network, thereby overlooking critical motion features in gesture actions.

To address this, we adopt a spatio-temporal graph modeling approach to capture finer gesture movement features. Here, action and audio features are represented as distinct graph structures, $\mathbf{G} = (v, e_1)$ and $\mathbf{R} = (v, e_2)$ respectively, where $v$ is the set of nodes (representing direction vectors for joint positions) and $e$ represents edges. Each node contains both coordinate and audio information. Drawing from previous findings ~\cite{van2016wavenet} that a single WaveNet can capture characteristics of diverse speakers by conditioning on speaker identity, we apply methods from ~\cite{wu2019graph} to capture hidden spatial dependencies and temporal information in gestures, while leveraging Wavenet to extract rhythmic cues from audio.

Our graph encoder comprises graph convolutional layers and 1D convolutional layers. In the graph convolutional layer, the model captures dependencies within the gesture feature space by initializing two learnable node embedding dictionaries, $\mathbf{E}_1,\mathbf{E}_2\in\mathbf{R}^{N\times e}$, forming an adaptive adjacency matrix.
\begin{equation}
{\mathbf{A}}_{\text{adapted}} = \text{SoftMax}\left(\text{ReLU}\left(\mathbf{E}_1 \odot \mathbf{E}_2^{T}\right)\right) \end{equation}
where $\mathbf{E}_1$ is the source node embedding and $\mathbf{E}_2$ is the target node embedding. Finally, our graph convolution layer is represented as follows:
\begin{align}
\mathbf{Z}_(r,g)^{(1:T)} = \sum_{j=0}^{J} & \, \mathbf{Q}_{f}^j \mathbf{[G, R]}^{(1:T)} \mathbf{W}_{j1} 
+ \mathbf{Q}_{b}^j \mathbf{[G, R]}^{(1:T)} \mathbf{W}_{j2} \notag \\ 
& + {\mathbf{A}}_{\text{adapted}}^j \mathbf{[G, R]}^{(1:T)} \mathbf{W}_{j3} \end{align}
where $\mathbf{Q}_f^J$ and $\mathbf{Q}_b^J$ respectively represent the power series of the forward and the backward transition matrix and $\mathbf{W}_{h}$ represents the model parameter matrix.

In the 1D convolutional layers, we employ dilated causal convolution ~\cite{yu2015multi} as our temporal convolution layer (TCN) to capture temporal trends and rhythmic features in audio. This method enables an exponentially large receptive field and efficiently handles long-range sequences, facilitating parallel computation and mitigating gradient explosion.

The dilated causal convolution preserves temporal order via zero-padding, ensuring predictions rely only on historical data. For a 1D input
$\mathbf{x}\in\mathbf{R}^{T}$ and a filter $\mathbf{f}\in\mathbf{R}^{H}$, the operation at time step \textit{t} is represented as \begin{equation}y_t=\sum_{i=0}^{K-1}f_ix_{t-d\cdot i}\end{equation} 
where $d$ is the dilation factor. Stacking layers with increasing dilation factors allows effective capture of rhythmic patterns while expanding the receptive field, enabling longer sequence processing with fewer layers and conserving resources.

\begin{figure}[htb]
	\centering
	\includegraphics[width=0.45\textwidth]{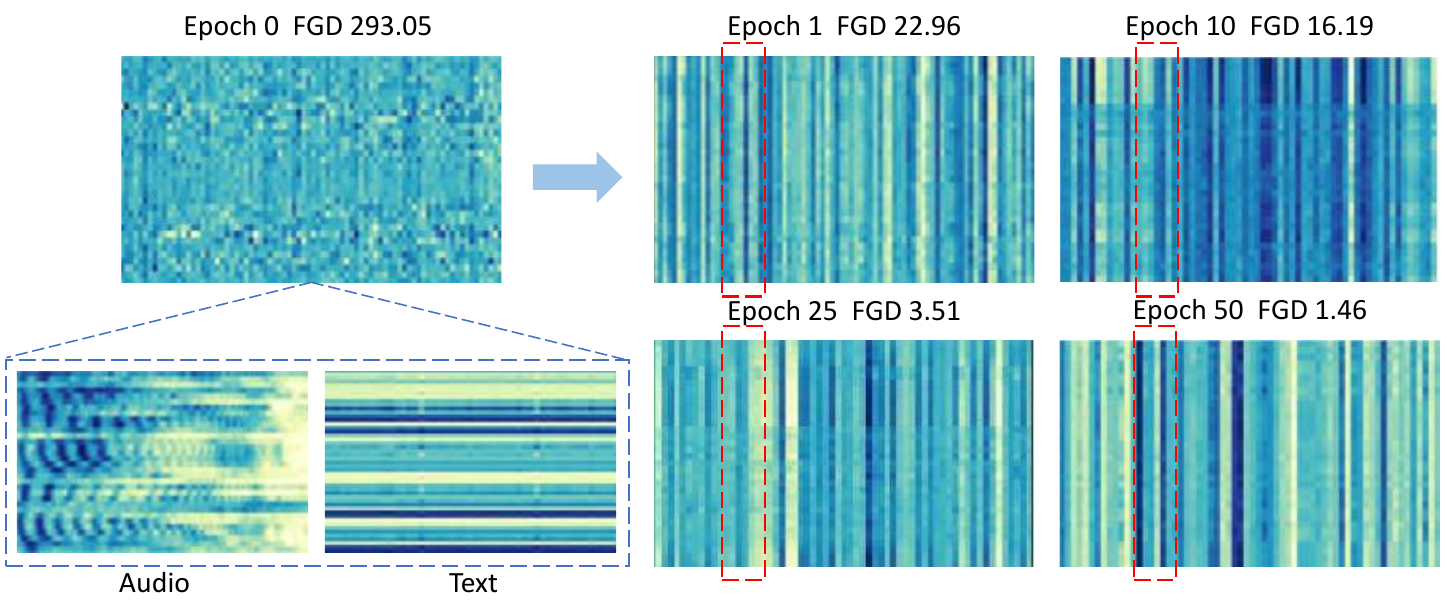}
    \vspace{-0.1in}
	\caption{\textbf{A showcase of reprogramming in audio-text cross-modality adaptation}. We visualize the features before and after reprogramming in the cross-modality adaptation, as well as the feature separation of audio and text before training. It is evident that the audio features are relatively noisier compared to the text features before training. After passing through the reprogramming layer, the correlation between audio and text increases as training progresses, showing a trend of alignment.}
\label{reprogram}
\vspace{-0.15in}
\end{figure}

\begin{figure*}[htb]
	\centering
	\includegraphics[width=\textwidth]{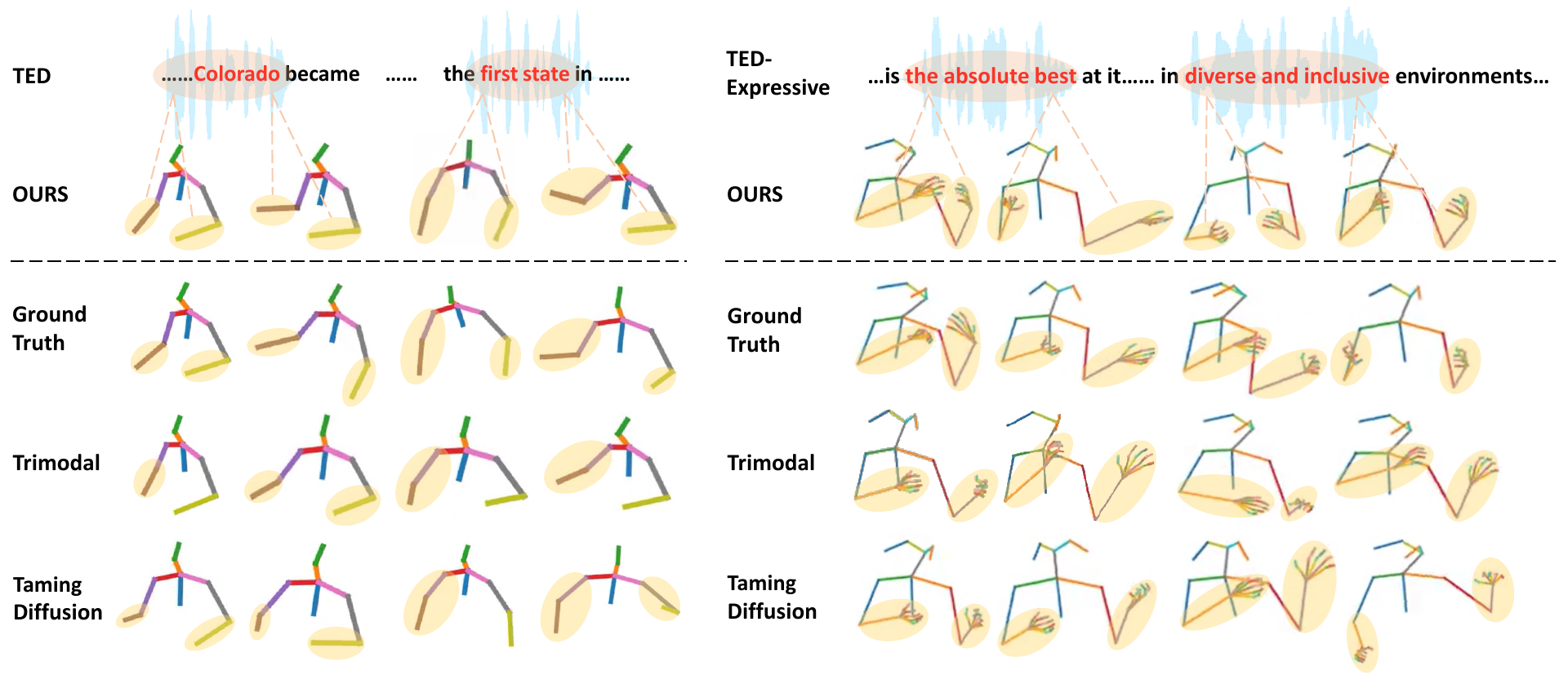}
    \vspace{-0.3in}
	\caption{\textbf{Visualization of generated gestures.} The gestures generated by our method more effectively capture the semantic information in the text, exhibiting a greater range of movement rhythm in the highlighted sections. We highlight the text and its corresponding gesture actions using red and yellow shading, respectively.}
    \label{result visible}
    \vspace{-0.2in}
\end{figure*}

\subsection{Human Pose Generation} 
\label{subsec:Generation}

\hspace{1em}For gesture generation, we use a multilayer bi-directional gated recurrent unit (GRU) network ~\cite{cho2014properties}. We use audio information as a bridge between semantic information and gesture-action information, and combine text, audio, and action features through topological fusion to form a multimodal feature $(\mathbf{Z}_(w,r)^{t}, \mathbf{Z}_(r,g)^{t})$ at each time point $t$. Also, we add the speaker's speech style features to the model learning. To train our model, the regression loss $\mathcal{L}_{\mathrm{Huber}}$ between the generated samples $\mathbf{g}$ and the groundtruth $\mathbf{\hat{g}}$ and the regression loss $\mathcal{L}_{\mathrm{style}}$ between the gestures $\mathbf{g}_{id}$ and $\mathbf{\hat{g}}_{id'}$ generated by different speakers are computed by smooth L1 loss , and the Kullback-Leibler divergence lkl is used to regularize the distributions of all the speakers on the embedding space to prevent the style embedding space from being too sparse .Finally we use the same method as in ~\cite{yoon2020speech} for adversarial training of our model. The overall learning objectives of the whole framework are as follows:
\begin{equation}
\begin{split}
\mathcal{L}_{\mathrm{gesture}} = & \ \alpha \cdot \mathcal{L}_{\mathrm{Huber}}(\mathbf{g}, \mathbf{\hat{g}}) + \beta \cdot \mathcal{L}_{\mathrm{style}}(\mathbf{g}_{id}, \mathbf{\hat{g}}_{id'}) \\
& + \gamma \cdot \mathcal{L}_{\mathrm{KLD}} + \lambda \cdot \mathcal{L}_{\mathrm{GAN}}
\end{split}
\end{equation}
where the $\alpha, \beta, \gamma, \lambda$ are weight coefficients.




\section{Experiments}

\begin{table*}[ht]
\centering
\begin{tabular}{lcccccc}
\toprule
& \multicolumn{3}{c}{TED Gesture~\cite{yoon2019robots,yoon2020speech}} & \multicolumn{3}{c}{TED Expressive~\cite{liu2022learning}} \\
\cmidrule(lr){2-4} \cmidrule(lr){5-7}
Methods & FGD↓ & BC↑ & Diversity↑ & FGD↓ & BC↑ & Diversity↑ \\
\midrule
Ground Truth & 0 & 0.698 & 108.525 & 0 & 0.703 & 178.827 \\
\midrule
Attention Seq2Seq~\cite{yoon2019robots} & 18.154 & 0.196 & 82.776 & 54.920 & 0.152 & 122.693 \\
Speech2Gesture~\cite{ginosar2019learning} & 19.254 & 0.668 & 93.802 & 54.650 & 0.679 & 142.489 \\
Joint Embedding~\cite{ahuja2019language2pose} & 22.083 & 0.200 & 90.138 & 64.555 & 0.130 & 120.627 \\
Trimodal~\cite{yoon2020speech} & 3.729 & 0.667 & 101.247 & 12.613 & 0.563 & 154.088 \\
HA2G~\cite{liu2022learning} & 3.072 & 0.672 & 104.322 & 5.306 & 0.641 & 173.899 \\
DiffGesture~\cite{zhu2023taming}  & \underline{1.506} & \underline{0.699} & \underline{106.722} & \underline{2.600} & \underline{0.718} & \underline{182.757} \\
\midrule
{HOP}(Ours)    & \textbf{1.406} & \textbf{0.762} & \textbf{108.176} & \textbf{1.815} & \textbf{0.738} & \textbf{183.332} \\
\bottomrule
\end{tabular}
\caption{\textbf{The Quantitative Results on TED Gesture ~\cite{yoon2019robots,yoon2020speech} and TED Expressive ~\cite{liu2022learning}.} We compare the proposed method  ~\cite{yoon2019robots,ginosar2019learning,ahuja2019language2pose,yoon2020speech,liu2022learning,zhu2023taming} based on topological fusion of heterogeneous multimodal learning with recent sota methods and ground truth. Lower FGD is better, higher BC and diversity are better.}
\label{quantitative results}
\vspace{-0.25in}
\end{table*}

\label{sec:experiments}

\noindent\textbf{TED Gesture.} The TED Gesture dataset~\cite{yoon2019robots,yoon2020speech} is a large-scale English-language dataset for speech-driven motion synthesis, comprising 1,766 TED talk videos by various speakers on diverse topics. It includes 3D human skeletons, aligned English transcripts, and audio data. Following the approach in ~\cite{yoon2020speech}, we resample human poses at 15 FPS and create input segments by sampling consecutive 34-frame windows with a stride of 10 frames, resulting in 252,109 segments totaling 106.1 hours. Each pose p  is represented by direction vectors of 10 upper body joints.

\noindent\textbf{TED Expressive.} Unlike TED Gesture, which includes only 10 upper body keypoints, the TED Expressive dataset~\cite{liu2022learning} provides richer detail by capturing both body and finger movements. Using the ExPose 3D pose estimator ~\cite{choutas2020monocular}, this dataset annotates the 3D coordinates of 43 keypoints, including 13 upper body joints and 30 finger joints.
\textbf{Baseline.} We compare our method with the state-of-the-art methods of recent years on two datasets. (1) Attention Seq2Seq ~\cite{yoon2019robots} generates gestures from speech text using the attention mechanism. (2) Speech2Gesture ~\cite{ginosar2019learning} takes the spectrogram of the audio as input and generates speech gestures by means of an encoder-decoder architecture and an adversarial training approach. (3) Joint Embedding ~\cite{ahuja2019language2pose} maps text and actions into the same embedding space and then generates gesture actions from the descriptive text of the actions. (4) Trimodal ~\cite{yoon2020speech} learns co-speech gestures from the trimodal context of text, audio and speaker identity, and its performance is a major breakthrough over the above methods. (5) HA2G ~\cite{liu2022learning} capturing information at different semantic granularities by extracting hierarchical audio features and then generating final gestures through hierarchical inference.(6)Taming Diffusion~\cite{zhu2023taming} is the state-of-the-art approach that establish the forward diffusion and the reverse conditional generation process process, processes multimodal data to generate gesture actions through a diffusion audio-gesture transformer.

\noindent\textbf{Implementation Details.} All seven methods above were experimented on the TED gesture dataset and the TED-Expressive dataset. For the text encoder we use the Bert-base-uncased model and do not need to update the parameters. The gesture actions are converted into a graph structure, where each direction vector represents a node and each node contains 3 dimensional features. The graph encoder structure we borrowed from ~\cite{wu2019graph} and resized the kernel size to . We used the Adam optimizerc $(lr=0.0001,\beta=(0.5,0.999))$  for 75 epochs of training, and all experiments were performed on a single NVIDIA RTX 6000 Ada.
\subsection{Quantitative Results}

\textbf{Evaluation Metrics. }We employ three metrics to evaluate the quality of results comprehensively. First, the Fréchet Gesture Distance (FGD) ~\cite{yoon2020speech} is utilized to measure the realism of the gesture movements. Second, the Beat Consistency (BC) ~\cite{liu2022learning} score is applied to assess the synchronization between co-speech gestures and the audio tempo, ensuring alignment with the rhythm of the accompanying speech. Lastly, we evaluate gesture diversity by calculating the average L1 distance between multiple generated body gestures ~\cite{liu2022learning}, reflecting the variability in gesture.

\noindent\textbf{Comparison with Baseline Models. }We evaluate our proposed method alongside several baseline models on two datasets, with the results summarized in Table~\ref{quantitative results}. While a higher BC score indicates denser motion beats, excessively high BC values can lead to overly frequent and unnatural gestures. Our method strikes a balance by generating gestures that not only maintain sufficient motion beats but also exhibit more natural movement. These results demonstrate that our approach effectively generates realistic gestures.

\noindent\textbf{Comparison of Model Generalization Capability. }The incorporation of topology fusion in our model enables enhanced preservation of structural information and the complex interrelations within the data. As illustrated in Table ~\ref{fewshot}, our model consistently outperforms the trimodal model~\cite{yoon2020speech} across all dataset proportions, with a more gradual and stable decline in performance as the dataset size decreases. This stability highlights the robustness, generalization, and adaptability of our method in handling multimodal data, particularly in scenarios with limited data availability.

\subsection{Qualitative Results}

\hspace{1em}We present visualizations of the gestures generated by our method in comparison to those produced by the Trimodal baseline method across both datasets, as shown in Figure ~\ref{result visible}. To facilitate analysis, we highlight focus words and their corresponding gestures with yellow and red shading, respectively. These visualizations reveal that gestures produced by our approach exhibit greater amplitude variation on selected focus words within sentences, effectively enhancing the conveyance of semantic information. Additionally, our method demonstrates rhythmic consistency and yields gestures that convey a more natural and realistic quality overall.


        


\noindent\textbf{User Study.} To evaluate the qualitative performance of the generated co-speech gestures, we conducted a user study involving 26 participants (13 females, 13 males) aged 18-30. Participants were asked to rate the quality, expressiveness and coherence of the motion in unlabeled video clips. A total of 30 cases were selected, including 20 from TED-Expressive and 10 from TED Gesture. For each case, participants viewed 8 videos (including ground truth), with the methods' order shuffled. The Mean Opinion Score (MOS) rating protocol was used, and participants assessed four aspects of the gestures: naturalness, smoothness, semantic and synchrony with speech. Ratings ranged from 1 to 5, with 5 indicating the best quality. As shown in Table~\ref{user_study}, our method received high ratings across all four criteria, indicating strong user approval of its performance.

\begin{table*}[htbp]
\centering
\begin{tabular}{lccccccc}
\toprule
Methods & Groundtruth & Seq2Seq & Joint Embedding & Trimodal & Attention Seq2Seq & HA2G & HOP(OURS) \\
\midrule
Naturalness & 4.16 & 1.36 & 1.52 & 3.66 & 2.88 & 3.13 & 3.92 \\
Smoothness & 3.97 & 4.48 & 4.32 & 3.87 & 2.23 & 2.92 & 3.77 \\
Semantic & 4.39 & 1.56 & 2.06 & 3.72 & 2.56 & 2.97 & 4.01 \\
Synchrony & 4.28 & 1.24 & 1.18 & 3.21 & 3.89 & 3.06 & 3.86 \\
\bottomrule
\end{tabular}
\caption{\textbf{User study results.} The ratings for motion naturalness, smoothness, semantic and synchrony, assessed on a scale from 1 to 5, with higher scores indicating better performance.}
\label{user_study}
\vspace{-0.1in}
\end{table*}


\begin{table*}[!h]
\centering
\begin{tabular}{lcccccc}
\toprule
& \multicolumn{3}{c}{HOP(Ours)} & \multicolumn{3}{c}{Trimodal [30]} \\
\cmidrule(lr){2-4} \cmidrule(lr){5-7}
Percentage of dataset & FGD↓ & BC↑ & Diversity↑ & FGD↓ & BC↑ & Diversity↑ \\
\midrule
Whole Dataset  & 1.406 & 0.762 & 108.176 & 3.729 & 0.667 & 101.247 \\
\midrule
90\% & 1.839 & 0.753 & 106.566 & 3.739 & 0.652 & 100.912 \\
80\% & 1.858 & 0.751 & 105.247 & 3.868 & 0.673 & 101.915 \\
70\% & 1.941 & 0.723 & 103.840 & 5.215 & 0.672 & 99.299 \\
60\% & 2.225 & 0.755 & 106.232 & 6.080 & 0.657 & 98.586 \\
50\% & 2.709 & 0.743 & 104.918 & 7.364 & 0.635 & 99.101 \\
\bottomrule
\end{tabular}
\caption{\textbf{Progressive learning results on varying percentages of TED training data.} All other training settings remain consistent with those in Table~\ref{quantitative results}. We gradually reduce the training dataset by 10\% increments to observe the model's performance changes under varying amounts of training data. When trained with 50\% of the data, our model still retains great learning effectiveness.}
\label{fewshot}
\vspace{-0.1in}
\end{table*}

\subsection{Ablation Studies}
\noindent\textbf{Text decoder. }To demonstrate the significance of incorporating a language model for semantic extraction, we conducted a series of ablation experiments: \textbf{1)} In the “w/o language model” variant, we replaced the language model with the method used by Trimodal~\cite{yoon2020speech} for extracting text features, omitting semantic analysis of the textual data. \textbf{2)} We separately employed BERT and GPT-2 to perform semantic analysis of the text. The results, presented in Table ~\ref{llm}, confirm the effectiveness of utilizing language models for semantic analysis within our framework.

\begin{table}[h]
    \centering
    \begin{tabular}{l|ccc}
    \hline
    \text{Methods} & \text{FGD}$\downarrow$ & \text{BC}$\uparrow$ & \text{Diversity}$\uparrow$ \\ \hline
    w/o Language Model & 1.955 & 0.701 & 105.311 \\ 
    GPT-2 & \textbf{1.319} & 0.753 & 107.036 \\ 
    \textbf{BERT} & 1.406  & \textbf{0.762}  & \textbf{108.176} \\ \hline
    \end{tabular}
    \caption{\textbf{Ablation study results of text decoder.} We investigate the performance of the proposed method without using a language model, as well as with different language models (including GPT-2 and BERT) as text encoders.}
    \label{llm}
\end{table}

\noindent\textbf{Reprogramming layers. }We analyze the reprogramming process for text and audio cross-modal adaptation, with experimental results shown in Figure~\ref{reprogram} . During the stochastic initialization phase, the text and audio data are initially scattered. However, as training progresses, the correlation between the two modalities improves, leading to better alignment of their features. This demonstrates the effectiveness of using reprogramming to align pairwise modal data in the gesture generation task.

\noindent\textbf{Model module.} This section presents an ablation study focused on two joint modules within our proposed model architecture. \textbf{1)} w/o Graph Encoder: In this configuration, we exclude the graph structure processing module, instead directly inputting the action data and audio features into the GRU. \textbf{2)} w/o Reprogramming Layer: Here, we remove the reprogramming module, omitting the extraction of semantic information latent within the audio. The results of these experiments, provided in Table ~\ref{w/o}, demonstrate the effectiveness of our model in leveraging multimodal data across heterogeneous topologies.

\begin{table}[h]
    \centering
    \begin{tabular}{l|ccc}
    \hline
    \text{Methods} & \text{FGD}$\downarrow$ & \text{BC}$\uparrow$ & \text{Diversity}$\uparrow$ \\ \hline
    w/o Graph Encoeder & 2.026 & 0.650 & 103.311 \\ 
    w/o Reprogramming Layer & 1.721 & 0.755 & 105.360 \\ 
    \textbf{Proposed (no ablation)} & \textbf{1.406}  & \textbf{0.762}  & \textbf{108.176} \\ \hline
    \end{tabular}
    \caption{\textbf{Ablation study on model modules.} We investigate the effectiveness of the proposed modules, Graph Encoder and Reprogramming Layer for cross-modality adaptation. The results demonstrate that these modules consistently enhance performance on the benchmarks.}
    \label{w/o}
        \vspace{-0.15in}
\end{table}



\section{Conclusion}
\hspace{1em}In this work, we present a novel approach \textbf{HOP} to co-speech gesture generation by explicitly modeling the interdependencies between gesture motion, audio rhythm, and text semantics. Unlike traditional methods that treat multimodal inputs as independent, our framework leverages the natural synchronization between audio and gesture, with audio rhythm acting as a crucial bridge to align gestures with both the temporal and semantic aspects of speech. Through the use of Mel spectrogram features and a spatiotemporal graph neural network, we achieve improved coherence and diversity in the generated gestures, surpassing existing methods in key performance metrics. Our approach, which incorporates reprogramming techniques for cross-modality adaptation, represents a significant step forward in the field of co-speech gesture generation. By capturing the intricate entanglements among text, audio and action, this work offers a new perspective for more smooth, natural and engaging human-agent interaction.



\begin{thebibliography}{54}
\providecommand{\natexlab}[1]{#1}
\providecommand{\url}[1]{\texttt{#1}}
\expandafter\ifx\csname urlstyle\endcsname\relax
  \providecommand{\doi}[1]{doi: #1}\else
  \providecommand{\doi}{doi: \begingroup \urlstyle{rm}\Url}\fi

\bibitem[Ahuja and Morency(2019)]{ahuja2019language2pose}
Chaitanya Ahuja and Louis-Philippe Morency.
\newblock Language2pose: Natural language grounded pose forecasting.
\newblock In \emph{2019 International Conference on 3D Vision (3DV)}, pages 719--728. IEEE, 2019.

\bibitem[Bhattacharya et~al.(2021)Bhattacharya, Childs, Rewkowski, and Manocha]{bhattacharya2021speech2affectivegestures}
Uttaran Bhattacharya, Elizabeth Childs, Nicholas Rewkowski, and Dinesh Manocha.
\newblock Speech2affectivegestures: Synthesizing co-speech gestures with generative adversarial affective expression learning.
\newblock In \emph{Proceedings of the 29th ACM International Conference on Multimedia}, pages 2027--2036, 2021.

\bibitem[Chen(2024)]{chen2024model}
Pin-Yu Chen.
\newblock Model reprogramming: Resource-efficient cross-domain machine learning.
\newblock In \emph{Proceedings of the AAAI Conference on Artificial Intelligence}, pages 22584--22591, 2024.

\bibitem[Chen et~al.(2021)Chen, Li, Yang, Li, and Liu]{chen2021multi}
Zhan Chen, Sicheng Li, Bing Yang, Qinghan Li, and Hong Liu.
\newblock Multi-scale spatial temporal graph convolutional network for skeleton-based action recognition.
\newblock In \emph{Proceedings of the AAAI conference on artificial intelligence}, pages 1113--1122, 2021.

\bibitem[Cho(2014)]{cho2014properties}
Kyunghyun Cho.
\newblock On the properties of neural machine translation: Encoder-decoder approaches.
\newblock \emph{arXiv preprint arXiv:1409.1259}, 2014.

\bibitem[Choutas et~al.(2020)Choutas, Pavlakos, Bolkart, Tzionas, and Black]{choutas2020monocular}
Vasileios Choutas, Georgios Pavlakos, Timo Bolkart, Dimitrios Tzionas, and Michael~J Black.
\newblock Monocular expressive body regression through body-driven attention.
\newblock In \emph{Computer Vision--ECCV 2020: 16th European Conference, Glasgow, UK, August 23--28, 2020, Proceedings, Part X 16}, pages 20--40. Springer, 2020.

\bibitem[Cui et~al.(2019)Cui, Cao, Pan, Zhang, and Wang]{cui2019deep}
Runpeng Cui, Zhong Cao, Weishen Pan, Changshui Zhang, and Jianqiang Wang.
\newblock Deep gesture video generation with learning on regions of interest.
\newblock \emph{IEEE Transactions on Multimedia}, 22\penalty0 (10):\penalty0 2551--2563, 2019.

\bibitem[Escalera et~al.(2017)Escalera, Athitsos, and Guyon]{escalera2017challenges}
Sergio Escalera, Vassilis Athitsos, and Isabelle Guyon.
\newblock Challenges in multi-modal gesture recognition.
\newblock \emph{Gesture recognition}, pages 1--60, 2017.

\bibitem[Ginosar et~al.(2019)Ginosar, Bar, Kohavi, Chan, Owens, and Malik]{ginosar2019learning}
Shiry Ginosar, Amir Bar, Gefen Kohavi, Caroline Chan, Andrew Owens, and Jitendra Malik.
\newblock Learning individual styles of conversational gesture.
\newblock In \emph{Proceedings of the IEEE/CVF Conference on Computer Vision and Pattern Recognition}, pages 3497--3506, 2019.

\bibitem[Goodfellow et~al.(2020)Goodfellow, Pouget-Abadie, Mirza, Xu, Warde-Farley, Ozair, Courville, and Bengio]{goodfellow2020generative}
Ian Goodfellow, Jean Pouget-Abadie, Mehdi Mirza, Bing Xu, David Warde-Farley, Sherjil Ozair, Aaron Courville, and Yoshua Bengio.
\newblock Generative adversarial networks.
\newblock \emph{Communications of the ACM}, 63\penalty0 (11):\penalty0 139--144, 2020.

\bibitem[Jin et~al.(2023)Jin, Wang, Ma, Chu, Zhang, Shi, Chen, Liang, Li, Pan, et~al.]{jin2023time}
Ming Jin, Shiyu Wang, Lintao Ma, Zhixuan Chu, James~Y Zhang, Xiaoming Shi, Pin-Yu Chen, Yuxuan Liang, Yuan-Fang Li, Shirui Pan, et~al.
\newblock Time-llm: Time series forecasting by reprogramming large language models.
\newblock \emph{arXiv preprint arXiv:2310.01728}, 2023.

\bibitem[Kalberer et~al.(2003)Kalberer, M{\"u}ller, and Gool]{kalberer2003visual}
Gregor~A Kalberer, Pascal M{\"u}ller, and Luc~Van Gool.
\newblock Visual speech, a trajectory in viseme space.
\newblock \emph{International journal of imaging systems and technology}, 13\penalty0 (1):\penalty0 74--84, 2003.

\bibitem[Kim et~al.(2023)Kim, Noh, Ham, and Ko]{kim2023mpe4g}
Gwantae Kim, Seonghyeok Noh, Insung Ham, and Hanseok Ko.
\newblock Mpe4g: Multimodal pretrained encoder for co-speech gesture generation.
\newblock In \emph{ICASSP 2023-2023 IEEE International Conference on Acoustics, Speech and Signal Processing (ICASSP)}, pages 1--5. IEEE, 2023.

\bibitem[Kipp et~al.(2007)Kipp, Neff, Kipp, and Albrecht]{kipp2007towards}
Michael Kipp, Michael Neff, Kerstin~H Kipp, and Irene Albrecht.
\newblock Towards natural gesture synthesis: Evaluating gesture units in a data-driven approach to gesture synthesis.
\newblock In \emph{Intelligent Virtual Agents: 7th International Conference, IVA 2007 Paris, France, September 17-19, 2007 Proceedings 7}, pages 15--28. Springer, 2007.

\bibitem[Kopp and Wachsmuth(2004)]{kopp2004synthesizing}
Stefan Kopp and Ipke Wachsmuth.
\newblock Synthesizing multimodal utterances for conversational agents.
\newblock \emph{Computer animation and virtual worlds}, 15\penalty0 (1):\penalty0 39--52, 2004.

\bibitem[Kucherenko et~al.(2023)Kucherenko, Nagy, Yoon, Woo, Nikolov, Tsakov, and Henter]{kucherenko2023genea}
Taras Kucherenko, Rajmund Nagy, Youngwoo Yoon, Jieyeon Woo, Teodor Nikolov, Mihail Tsakov, and Gustav~Eje Henter.
\newblock The genea challenge 2023: A large-scale evaluation of gesture generation models in monadic and dyadic settings.
\newblock In \emph{Proceedings of the 25th International Conference on Multimodal Interaction}, pages 792--801, 2023.

\bibitem[Levine et~al.(2010)Levine, Kr{\"a}henb{\"u}hl, Thrun, and Koltun]{levine2010gesture}
Sergey Levine, Philipp Kr{\"a}henb{\"u}hl, Sebastian Thrun, and Vladlen Koltun.
\newblock Gesture controllers.
\newblock In \emph{Acm siggraph 2010 papers}, pages 1--11. 2010.

\bibitem[Li et~al.(2021)Li, Kang, Pei, Zhe, Zhang, He, and Bao]{li2021audio2gestures}
Jing Li, Di Kang, Wenjie Pei, Xuefei Zhe, Ying Zhang, Zhenyu He, and Linchao Bao.
\newblock Audio2gestures: Generating diverse gestures from speech audio with conditional variational autoencoders.
\newblock In \emph{Proceedings of the IEEE/CVF International Conference on Computer Vision}, pages 11293--11302, 2021.

\bibitem[Liu et~al.(2022{\natexlab{a}})Liu, Zhu, Iwamoto, Peng, Li, Zhou, Bozkurt, and Zheng]{liu2022beat}
Haiyang Liu, Zihao Zhu, Naoya Iwamoto, Yichen Peng, Zhengqing Li, You Zhou, Elif Bozkurt, and Bo Zheng.
\newblock Beat: A large-scale semantic and emotional multi-modal dataset for conversational gestures synthesis.
\newblock In \emph{European conference on computer vision}, pages 612--630. Springer, 2022{\natexlab{a}}.

\bibitem[Liu et~al.(2024)Liu, Zhu, Becherini, Peng, Su, Zhou, Zhe, Iwamoto, Zheng, and Black]{liu2024emage}
Haiyang Liu, Zihao Zhu, Giorgio Becherini, Yichen Peng, Mingyang Su, You Zhou, Xuefei Zhe, Naoya Iwamoto, Bo Zheng, and Michael~J Black.
\newblock Emage: Towards unified holistic co-speech gesture generation via expressive masked audio gesture modeling.
\newblock In \emph{Proceedings of the IEEE/CVF Conference on Computer Vision and Pattern Recognition}, pages 1144--1154, 2024.

\bibitem[Liu et~al.(2022{\natexlab{b}})Liu, Wu, Zhou, Du, Wu, Lin, and Liu]{liu2022audio}
Xian Liu, Qianyi Wu, Hang Zhou, Yuanqi Du, Wayne Wu, Dahua Lin, and Ziwei Liu.
\newblock Audio-driven co-speech gesture video generation.
\newblock \emph{Advances in Neural Information Processing Systems}, 35:\penalty0 21386--21399, 2022{\natexlab{b}}.

\bibitem[Liu et~al.(2022{\natexlab{c}})Liu, Wu, Zhou, Xu, Qian, Lin, Zhou, Wu, Dai, and Zhou]{liu2022learning}
Xian Liu, Qianyi Wu, Hang Zhou, Yinghao Xu, Rui Qian, Xinyi Lin, Xiaowei Zhou, Wayne Wu, Bo Dai, and Bolei Zhou.
\newblock Learning hierarchical cross-modal association for co-speech gesture generation.
\newblock In \emph{Proceedings of the IEEE/CVF Conference on Computer Vision and Pattern Recognition}, pages 10462--10472, 2022{\natexlab{c}}.

\bibitem[Liu et~al.(2021)Liu, Mohammadi, Song, and Johal]{liu2021speech}
Yu Liu, Gelareh Mohammadi, Yang Song, and Wafa Johal.
\newblock Speech-based gesture generation for robots and embodied agents: A scoping review.
\newblock In \emph{Proceedings of the 9th International Conference on Human-Agent Interaction}, pages 31--38, 2021.

\bibitem[Liu et~al.(2023)Liu, Magliacane, Kofinas, and Gavves]{liu2023graph}
Yongtuo Liu, Sara Magliacane, Miltiadis Kofinas, and Efstratios Gavves.
\newblock Graph switching dynamical systems.
\newblock In \emph{International Conference on Machine Learning}, pages 21867--21883. PMLR, 2023.

\bibitem[Loehr(2012)]{loehr2012temporal}
Daniel~P Loehr.
\newblock Temporal, structural, and pragmatic synchrony between intonation and gesture.
\newblock \emph{Laboratory phonology}, 3\penalty0 (1):\penalty0 71--89, 2012.

\bibitem[Marsella et~al.(2013)Marsella, Xu, Lhommet, Feng, Scherer, and Shapiro]{marsella2013virtual}
Stacy Marsella, Yuyu Xu, Margaux Lhommet, Andrew Feng, Stefan Scherer, and Ari Shapiro.
\newblock Virtual character performance from speech.
\newblock In \emph{Proceedings of the 12th ACM SIGGRAPH/Eurographics symposium on computer animation}, pages 25--35, 2013.

\bibitem[Mughal et~al.(2024)Mughal, Dabral, Habibie, Donatelli, Habermann, and Theobalt]{mughal2024convofusion}
Muhammad~Hamza Mughal, Rishabh Dabral, Ikhsanul Habibie, Lucia Donatelli, Marc Habermann, and Christian Theobalt.
\newblock Convofusion: Multi-modal conversational diffusion for co-speech gesture synthesis.
\newblock In \emph{Proceedings of the IEEE/CVF Conference on Computer Vision and Pattern Recognition}, pages 1388--1398, 2024.

\bibitem[Nagrani et~al.(2021)Nagrani, Yang, Arnab, Jansen, Schmid, and Sun]{nagrani2021attention}
Arsha Nagrani, Shan Yang, Anurag Arnab, Aren Jansen, Cordelia Schmid, and Chen Sun.
\newblock Attention bottlenecks for multimodal fusion.
\newblock \emph{Advances in neural information processing systems}, 34:\penalty0 14200--14213, 2021.

\bibitem[Neff et~al.(2008)Neff, Kipp, Albrecht, and Seidel]{neff2008gesture}
Michael Neff, Michael Kipp, Irene Albrecht, and Hans-Peter Seidel.
\newblock Gesture modeling and animation based on a probabilistic re-creation of speaker style.
\newblock \emph{ACM Transactions On Graphics (TOG)}, 27\penalty0 (1):\penalty0 1--24, 2008.

\bibitem[Nyatsanga et~al.(2023)Nyatsanga, Kucherenko, Ahuja, Henter, and Neff]{nyatsanga2023comprehensive}
Simbarashe Nyatsanga, Taras Kucherenko, Chaitanya Ahuja, Gustav~Eje Henter, and Michael Neff.
\newblock A comprehensive review of data-driven co-speech gesture generation.
\newblock In \emph{Computer Graphics Forum}, pages 569--596. Wiley Online Library, 2023.

\bibitem[Owens and Blazek(1985)]{owens1985visemes}
Elmer Owens and Barbara Blazek.
\newblock Visemes observed by hearing-impaired and normal-hearing adult viewers.
\newblock \emph{Journal of Speech, Language, and Hearing Research}, 28\penalty0 (3):\penalty0 381--393, 1985.

\bibitem[Piechocki et~al.(2023)Piechocki, Wang, and Bocus]{piechocki2023multimodal}
Robert~J Piechocki, Xiaoyang Wang, and Mohammud~J Bocus.
\newblock Multimodal sensor fusion in the latent representation space.
\newblock \emph{Scientific Reports}, 13\penalty0 (1):\penalty0 2005, 2023.

\bibitem[Poggi et~al.(2005)Poggi, Pelachaud, de~Rosis, Carofiglio, and De~Carolis]{poggi2005greta}
Isabella Poggi, Catherine Pelachaud, Fiorella de Rosis, Valeria Carofiglio, and Berardina De~Carolis.
\newblock Greta. a believable embodied conversational agent.
\newblock In \emph{Multimodal intelligent information presentation}, pages 3--25. Springer, 2005.

\bibitem[Ramachandram and Taylor(2017)]{ramachandram2017deep}
Dhanesh Ramachandram and Graham~W Taylor.
\newblock Deep multimodal learning: A survey on recent advances and trends.
\newblock \emph{IEEE signal processing magazine}, 34\penalty0 (6):\penalty0 96--108, 2017.

\bibitem[Sheng et~al.(2019)Sheng, Huang, and Pavlovskiy]{sheng2019high}
Leyuan Sheng, Dong-Yan Huang, and Evgeniy~N Pavlovskiy.
\newblock High-quality speech synthesis using super-resolution mel-spectrogram.
\newblock \emph{arXiv preprint arXiv:1912.01167}, 2019.

\bibitem[Shi et~al.(2024{\natexlab{a}})Shi, Luo, Song, Li, and Pan]{shi2024deep}
Guangsi Shi, Linhao Luo, Yongze Song, Jing Li, and Shirui Pan.
\newblock Deep transformer-based heterogeneous spatiotemporal graph learning for geographical traffic forecasting.
\newblock \emph{iScience}, 27\penalty0 (7), 2024{\natexlab{a}}.

\bibitem[Shi et~al.(2024{\natexlab{b}})Shi, Zhang, Jin, Pan, and Philip]{shi2024towards}
Guangsi Shi, Daokun Zhang, Ming Jin, Shirui Pan, and S~Yu Philip.
\newblock Towards complex dynamic physics system simulation with graph neural ordinary equations.
\newblock \emph{Neural Networks}, 176:\penalty0 106341, 2024{\natexlab{b}}.

\bibitem[Su et~al.(2019)Su, Zhao, Niu, Liu, Sun, and Pei]{su2019robust}
Ya Su, Youjian Zhao, Chenhao Niu, Rong Liu, Wei Sun, and Dan Pei.
\newblock Robust anomaly detection for multivariate time series through stochastic recurrent neural network.
\newblock In \emph{Proceedings of the 25th ACM SIGKDD international conference on knowledge discovery \& data mining}, pages 2828--2837, 2019.

\bibitem[Teshima et~al.(2023)Teshima, Wake, Thomas, Nakashima, Kawasaki, and Ikeuchi]{teshima2023act2g}
Hitoshi Teshima, Naoki Wake, Diego Thomas, Yuta Nakashima, Hiroshi Kawasaki, and Katsushi Ikeuchi.
\newblock Act2g: Attention-based contrastive learning for text-to-gesture generation.
\newblock \emph{Proceedings of the ACM on Computer Graphics and Interactive Techniques}, 6\penalty0 (3):\penalty0 1--17, 2023.

\bibitem[Van Den~Oord et~al.(2016)Van Den~Oord, Dieleman, Zen, Simonyan, Vinyals, Graves, Kalchbrenner, Senior, Kavukcuoglu, et~al.]{van2016wavenet}
Aaron Van Den~Oord, Sander Dieleman, Heiga Zen, Karen Simonyan, Oriol Vinyals, Alex Graves, Nal Kalchbrenner, Andrew Senior, Koray Kavukcuoglu, et~al.
\newblock Wavenet: A generative model for raw audio.
\newblock \emph{arXiv preprint arXiv:1609.03499}, 12, 2016.

\bibitem[Wagner et~al.(2014)Wagner, Malisz, and Kopp]{wagner2014gesture}
Petra Wagner, Zofia Malisz, and Stefan Kopp.
\newblock Gesture and speech in interaction: An overview, 2014.

\bibitem[Wang et~al.(2024)Wang, Zhang, Ye, Zeng, and Mei]{wang2024cross}
Zheng Wang, Wei Zhang, Long Ye, Dan Zeng, and Tao Mei.
\newblock Cross-modal quantization for co-speech gesture generation.
\newblock \emph{IEEE Transactions on Multimedia}, 2024.

\bibitem[Wei et~al.(2020)Wei, Zhang, Li, Zhang, and Wu]{wei2020multi}
Xi Wei, Tianzhu Zhang, Yan Li, Yongdong Zhang, and Feng Wu.
\newblock Multi-modality cross attention network for image and sentence matching.
\newblock In \emph{Proceedings of the IEEE/CVF conference on computer vision and pattern recognition}, pages 10941--10950, 2020.

\bibitem[Wolfert et~al.(2022)Wolfert, Robinson, and Belpaeme]{wolfert2022review}
Pieter Wolfert, Nicole Robinson, and Tony Belpaeme.
\newblock A review of evaluation practices of gesture generation in embodied conversational agents.
\newblock \emph{IEEE Transactions on Human-Machine Systems}, 52\penalty0 (3):\penalty0 379--389, 2022.

\bibitem[Wu et~al.(2019)Wu, Pan, Long, Jiang, and Zhang]{wu2019graph}
Zonghan Wu, Shirui Pan, Guodong Long, Jing Jiang, and Chengqi Zhang.
\newblock Graph wavenet for deep spatial-temporal graph modeling.
\newblock In \emph{The 28th International Joint Conference on Artificial Intelligence (IJCAI)}, 2019.

\bibitem[Yang et~al.(2021)Yang, Tsai, and Chen]{yang2021voice2series}
Chao-Han~Huck Yang, Yun-Yun Tsai, and Pin-Yu Chen.
\newblock Voice2series: Reprogramming acoustic models for time series classification.
\newblock In \emph{International conference on machine learning}, pages 11808--11819. PMLR, 2021.

\bibitem[Yang et~al.(2023)Yang, Wu, Li, Zhang, Hao, Bao, Cheng, and Xiao]{yang2023diffusestylegesture}
Sicheng Yang, Zhiyong Wu, Minglei Li, Zhensong Zhang, Lei Hao, Weihong Bao, Ming Cheng, and Long Xiao.
\newblock Diffusestylegesture: Stylized audio-driven co-speech gesture generation with diffusion models.
\newblock \emph{arXiv preprint arXiv:2305.04919}, 2023.

\bibitem[Yin et~al.(2023)Yin, Fu, Zhao, Li, Sun, Xu, and Chen]{yin2023survey}
Shukang Yin, Chaoyou Fu, Sirui Zhao, Ke Li, Xing Sun, Tong Xu, and Enhong Chen.
\newblock A survey on multimodal large language models.
\newblock \emph{arXiv preprint arXiv:2306.13549}, 2023.

\bibitem[Yoon et~al.(2019)Yoon, Ko, Jang, Lee, Kim, and Lee]{yoon2019robots}
Youngwoo Yoon, Woo-Ri Ko, Minsu Jang, Jaeyeon Lee, Jaehong Kim, and Geehyuk Lee.
\newblock Robots learn social skills: End-to-end learning of co-speech gesture generation for humanoid robots.
\newblock In \emph{2019 International Conference on Robotics and Automation (ICRA)}, pages 4303--4309. IEEE, 2019.

\bibitem[Yoon et~al.(2020)Yoon, Cha, Lee, Jang, Lee, Kim, and Lee]{yoon2020speech}
Youngwoo Yoon, Bok Cha, Joo-Haeng Lee, Minsu Jang, Jaeyeon Lee, Jaehong Kim, and Geehyuk Lee.
\newblock Speech gesture generation from the trimodal context of text, audio, and speaker identity.
\newblock \emph{ACM Transactions on Graphics (TOG)}, 39\penalty0 (6):\penalty0 1--16, 2020.

\bibitem[Yoon et~al.(2022)Yoon, Wolfert, Kucherenko, Viegas, Nikolov, Tsakov, and Henter]{yoon2022genea}
Youngwoo Yoon, Pieter Wolfert, Taras Kucherenko, Carla Viegas, Teodor Nikolov, Mihail Tsakov, and Gustav~Eje Henter.
\newblock The genea challenge 2022: A large evaluation of data-driven co-speech gesture generation.
\newblock In \emph{Proceedings of the 2022 International Conference on Multimodal Interaction}, pages 736--747, 2022.

\bibitem[Yu(2015)]{yu2015multi}
F Yu.
\newblock Multi-scale context aggregation by dilated convolutions.
\newblock \emph{arXiv preprint arXiv:1511.07122}, 2015.

\bibitem[Yuan et~al.(2021)Yuan, Lin, Kuen, Zhang, Wang, Maire, Kale, and Faieta]{yuan2021multimodal}
Xin Yuan, Zhe Lin, Jason Kuen, Jianming Zhang, Yilin Wang, Michael Maire, Ajinkya Kale, and Baldo Faieta.
\newblock Multimodal contrastive training for visual representation learning.
\newblock In \emph{Proceedings of the IEEE/CVF Conference on Computer Vision and Pattern Recognition}, pages 6995--7004, 2021.

\bibitem[Zhu et~al.(2023)Zhu, Liu, Liu, Qian, Liu, and Yu]{zhu2023taming}
Lingting Zhu, Xian Liu, Xuanyu Liu, Rui Qian, Ziwei Liu, and Lequan Yu.
\newblock Taming diffusion models for audio-driven co-speech gesture generation.
\newblock In \emph{Proceedings of the IEEE/CVF Conference on Computer Vision and Pattern Recognition}, pages 10544--10553, 2023.

\end{thebibliography}

\clearpage
\appendix
\clearpage
\setcounter{page}{1}
\section{More Details of Audio-Action Cross-modality Adaptation}
\hspace{1em}The process of cross-modal adaptation between audio data and action data is illustrated in Fig.~\ref{audio-action}. Both modalities are transformed into a spatio-temporal graph structure and subsequently processed by the graph encoder ~\cite{wu2019graph}. The resulting fused features encapsulate action-related characteristics derived from gestures and rhythmic attributes extracted from the audio signals. Notably, the action features are segmented longitudinally into four parts, indicating that temporal action features with a time step of 4 have been effectively extracted from the action data.

In Fig.~\ref{graph}, we further analyze the role of the adaptive neighborhood matrix within the spatio-temporal graph encoder. To enhance the clarity of feature representation in the adaptive neighborhood matrix, we utilize the TED-Expressive datasets ~\cite{liu2022learning}. This dataset is particularly suitable as it includes a larger number of joints, each represented as nodes containing 3D joint features. The visualization reveals that certain columns in the matrix exhibit a higher density of high-value points, indicating that some nodes exert a stronger influence on other nodes, whereas others exhibit weaker interactions. For instance, column a displays a significantly higher concentration of high-value points compared to column b. This observation suggests that the joint action at node a likely represents a latent structural feature inherent in the action data.

\begin{figure}[h]
	\centering
        \includegraphics[width=0.45\textwidth]{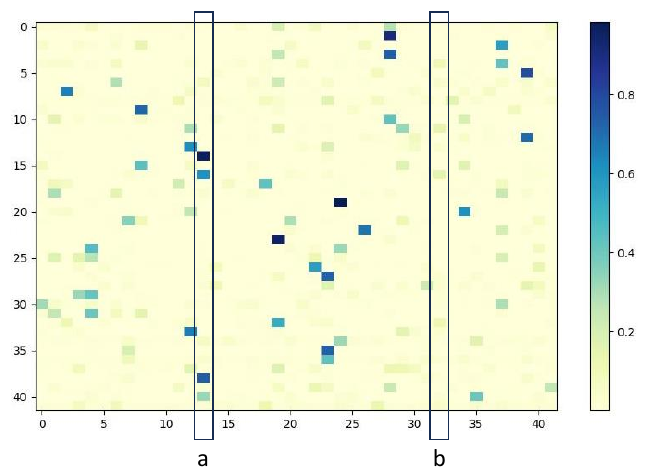}
    \vspace{-0.1in}
	\caption{\textbf{The visualization of adaptive adjacency matrix.}}
    \label{graph}
    \vspace{-0.2in}
\end{figure}

\begin{figure*}[htb]
	\centering
	\includegraphics[width=\textwidth]{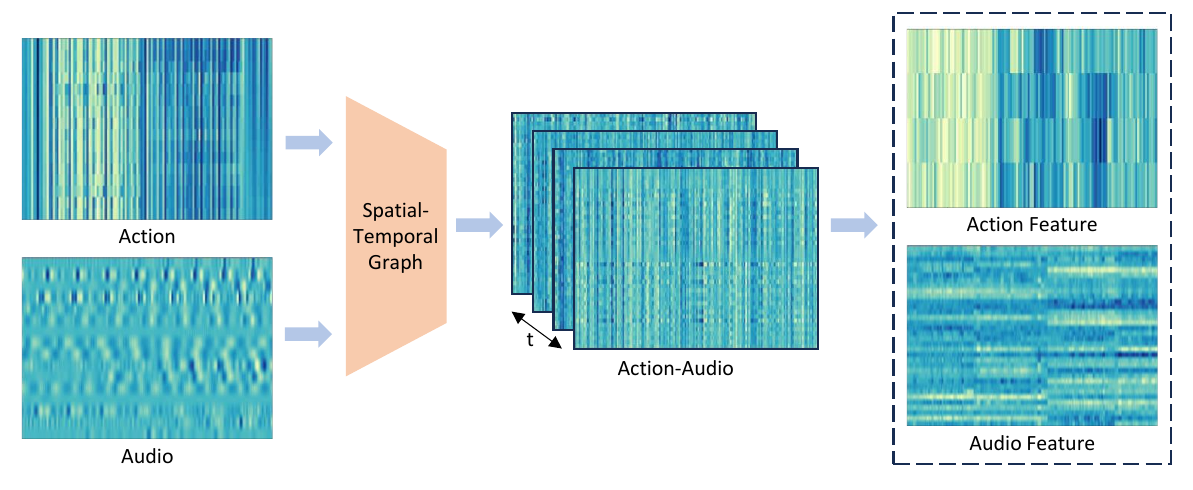}
    \vspace{-0.3in}
	\caption{\textbf{Demonstration of the Audio-Action cross-modality adaptation process}. Audio and text data are cross-modally adapted using a spatial-temporal graph encoder~\cite{wu2019graph}, enabling the fusion of cross-modal features that incorporate action features and rhythmic characteristics from the audio.}
    \label{audio-action}
    \vspace{-0.2in}
\end{figure*}

\section{Comparison with Baselines}
\hspace{1em}The visualized results of gestures generated by our method, alongside several baseline methods, are presented in Fig.~\ref{all_result}. These visualizations illustrate the diversity of gestures produced on the TED dataset, with instances of overly sparse gesture actions highlighted in red. While the gestures generated by our approach are comparable to those of the method proposed in ~\cite{yoon2020speech, liu2022learning,zhu2023taming}, in terms of BC and diversity metrics, the diversity visualization reveals that our method produces gestures with a more pronounced sense of rhythmic movement. Furthermore, compared to the method proposed in ~\cite{yoon2019robots,ginosar2019learning,ahuja2019language2pose}, the gestures generated by our approach exhibit greater vividness and convey richer semantic information.

\begin{figure*}[htb]
	\centering
	\includegraphics[width=\textwidth]{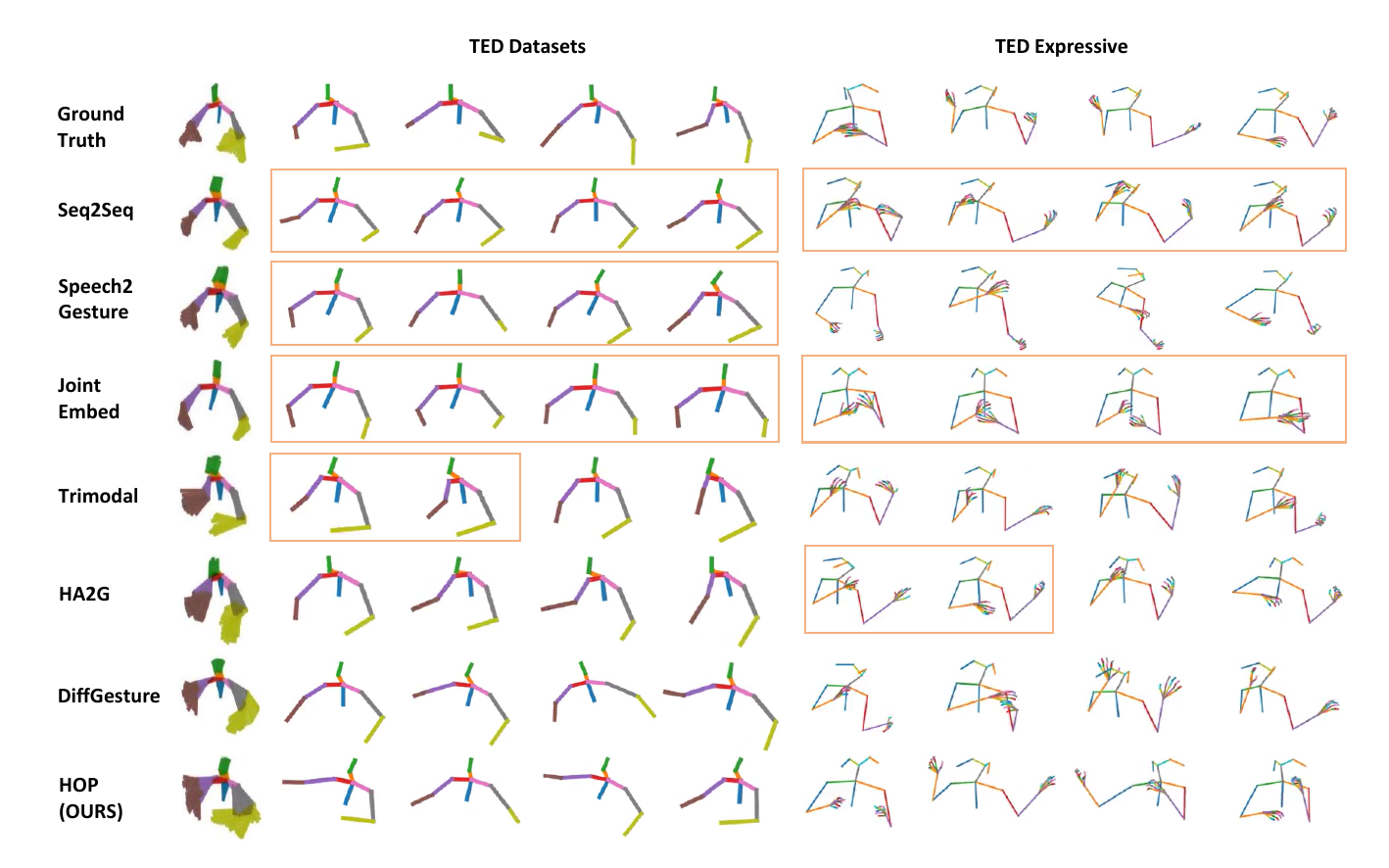}
    \vspace{-0.3in}
	\caption{\textbf{Visualization of generated gestures.} We compared the gesture visualization results on these two datasets with those generated by BASELINE ~\cite{yoon2019robots,ginosar2019learning,ahuja2019language2pose,yoon2020speech, liu2022learning,zhu2023taming}. Additionally, we present the visualization results of gesture diversity on the TED dataset, which clearly demonstrate that our approach significantly outperforms other methods in gesture diversity. Results with an insufficient number of gestures are highlighted with a red box.}
    \label{all_result}
    \vspace{-0.2in}
\end{figure*}

\section{More Details of HOP Model}
\textbf{Reprogramming Module. }Mel-spectral features were extracted from the raw audio and transformed into a format compatible with the input space of the large language model (LLM) through a reprogramming methodology. The core of the reprogramming module ~\cite{jin2023time} is the cross-attention mechanism, which facilitates the alignment and integration of audio features into the LLM framework. The detailed operations within the cross-attention mechanism, along with the corresponding output feature dimensions, are outlined in Table ~\ref{reprogramming detail}.

\noindent \textbf{Reprogramming Process. }The reprogramming module ~\cite{jin2023time} was first applied to the gesture generation task. To provide a more detailed explanation of the workflow and underlying mechanism of the reprogramming module, the corresponding algorithm is presented in Algorithm~\ref{reprogramming}.
\begin{algorithm}[h]
\DontPrintSemicolon 
\caption{Reprogramming Layer Forward Pass}
\label{reprogramming}
\KwIn{$target\_embedding \in \mathbb{R}^{B \times L \times d}$, $source\_embedding \in \mathbb{R}^{S \times d\_llm}$, $value\_embedding \in \mathbb{R}^{S \times d\_llm}$}
\KwOut{Reprogrammed output embedding $output \in \mathbb{R}^{B \times L \times d\_llm}$}
\SetAlgoLined
\textbf{Initialization:} \\
\Indp
Initialize $W_q, W_k, W_v, W_{out}$ for query, key, value, and output projections \\
\Indm
Reshape $target\_embedding$: \\
\Indp
$Q \gets W_q \cdot target\_embedding \to \mathbb{R}^{B \times L \times H \times -1}$ \\
\Indm
Reshape $source\_embedding$: \\
\Indp
$K \gets W_k \cdot source\_embedding \to \mathbb{R}^{S \times H \times -1}$ \\
\Indm
Reshape $value\_embedding$: \\
\Indp
$V \gets W_v \cdot value\_embedding \to \mathbb{R}^{S \times H \times -1}$ \\
\Indm

\textbf{Reprogramming:} \\
\Indp
Compute scaling factor: $scale \gets 1 / \sqrt{E}$ \\
Compute attention scores: $scores \gets Q \cdot K^\top$ \\
Apply Softmax and Dropout: $A \gets \text{Softmax}(scores \cdot scale)$ \\
Compute reprogrammed embedding: $reprogrammed\_embedding \gets A \cdot V$ \\
\Indm

Reshape $reprogrammed\_embedding$ to $(B, L, -1)$ \\
Apply ReLU activation: $output \gets \text{ReLU}(reprogrammed\_embedding)$ \\
Final projection: $output \gets W_{out} \cdot output$ \\
\Return $output$
\end{algorithm}

\noindent\textbf{Spatial-Temporal Graph Encoder. }The original audio data was processed using a sliding window approach, with a window size of 3400 and a step size of 2191. The audio data was then converted into a graph structure by increasing its dimensionality. Simultaneously, the action data was represented as a graph, with the number of joints serving as the nodes, each containing the 3D features of the respective joints. These graph representations of audio and action data were processed using Graph WaveNet, enabling the model to learn the rhythmic features from the audio and the action features from the gestures. The detailed operations and corresponding output feature dimensions are presented in Table ~\ref{graph encoder detail}.

\begin{table}[h]
    \centering
    \begin{tabular}{l|c}
    \hline
    \text{Operations} & \text{Feature Map Shapes} \\ \hline
    Input Mel-Spectral & $256\times34\times 128$ \\ 
    Input word embeddings & $1500 \times 768$ \\ 
    Query Linear(128,1024) & $256 \times 34 \times 1024$ \\
    Key Linear(768,1024) & $1500 \times 1024$ \\
    Value Linear(768,1024) & $1500 \times 1024$ \\
    Out Linear(1024,768) & $256 \times 34 \times 768$ \\ \hline
    \end{tabular}
    \caption{\textbf{Detailed feature Shapes in reprogramming module.} }
    \label{reprogramming detail}
        \vspace{-0.15in}
\end{table}

\begin{table}[h]
    \centering
    \begin{tabular}{l|c}
    \hline
    \text{Operations} & \text{Feature Map Shapes} \\ \hline
    Input Audio & $1\times36267$ \\ 
    Input Action & $256\times16\times27$ \\ 
    Audio Matrix Converter & $256\times16\times9\times170$ \\
    Action Matrix Converter & $256\times16\times9\times 3$ \\
    Graph Wavenet & $256\times173\times9\times 4$ \\ \hline
    \end{tabular}
    \caption{\textbf{Detailed feature shapes in the spatial-temporal graph encoder.} }
    \label{graph encoder detail}
    \vspace{-0.15in}
\end{table}

\end{document}